%% file: acl_latex.tex
\pdfoutput=1

\documentclass[11pt]{article}
\usepackage{enumitem}
\usepackage{adjustbox}

\usepackage{subcaption} 
\usepackage{afterpage}

\usepackage[final]{acl}
\usepackage{subcaption}

\usepackage{amsmath}
\usepackage{amsmath,amssymb}

\usepackage[nameinlink,noabbrev]{cleveref}

\usepackage{times}
\usepackage{latexsym}

\usepackage[T1]{fontenc}

\usepackage[utf8]{inputenc}

\usepackage{microtype}

\usepackage{inconsolata}

\usepackage{array}
\usepackage{graphicx}
\usepackage{booktabs}
\usepackage{tabularx}
\usepackage{multirow}

\usepackage[table]{xcolor}
\definecolor{pale_green}{HTML}{ECF5E4}
\definecolor{mplPurple}{HTML}{9467BD}

\usepackage{multirow,makecell}

\usepackage{fancyvrb}
\title{SyncThink: A Training-Free Strategy to Align Inference Termination with Reasoning Saturation}

\author{
Gengyang Li$^{1,2}$\thanks{~~Equal contribution.} \quad 
Wang Cai$^{1,2}$\footnotemark[1] \quad 
Yifeng Gao $^{1,2}$ \quad
Yunfang Wu$^{1,3}$\thanks{~~Corresponding author.} \\
$^1$National Key Laboratory for Multimedia Information Processing, Peking University \\
$^2$School of Software and Microelectronics, Peking University \\
$^3$School of Computer Science, Peking University \\
\texttt{\{ligengyang, caiwang, yifgao26\}@stu.pku.edu.cn} \quad
\texttt{wuyf}@pku.edu.cn 
}

\begin{document}
\maketitle
\begin{abstract}

Chain-of-Thought (CoT) prompting improves reasoning but often produces long and redundant traces that substantially increase inference cost. 
We present \textbf{SyncThink}, a training-free and plug-and-play decoding method that reduces CoT overhead without modifying model weights. 
We find that answer tokens attend weakly to early reasoning and focus on \texttt{</think>}, indicating an information bottleneck.
Building on this observation, SyncThink monitors the model’s own reasoning-transition signal and terminates reasoning.
Experiments on GSM8K, MMLU, GPQA, and BBH across three DeepSeek-R1 distilled models show that SyncThink achieves 62.00\% average Top@1 accuracy using 656 generated tokens and 28.68s latency, compared to 61.22\%, 2141 tokens, and 92.01s for full CoT decoding. On long-horizon tasks such as GPQA, SyncThink can further yield up to +8.1 absolute accuracy by preventing over-thinking.

\end{abstract}

\section{Introduction}

\begin{figure}[!t]
  \includegraphics[width=\linewidth]{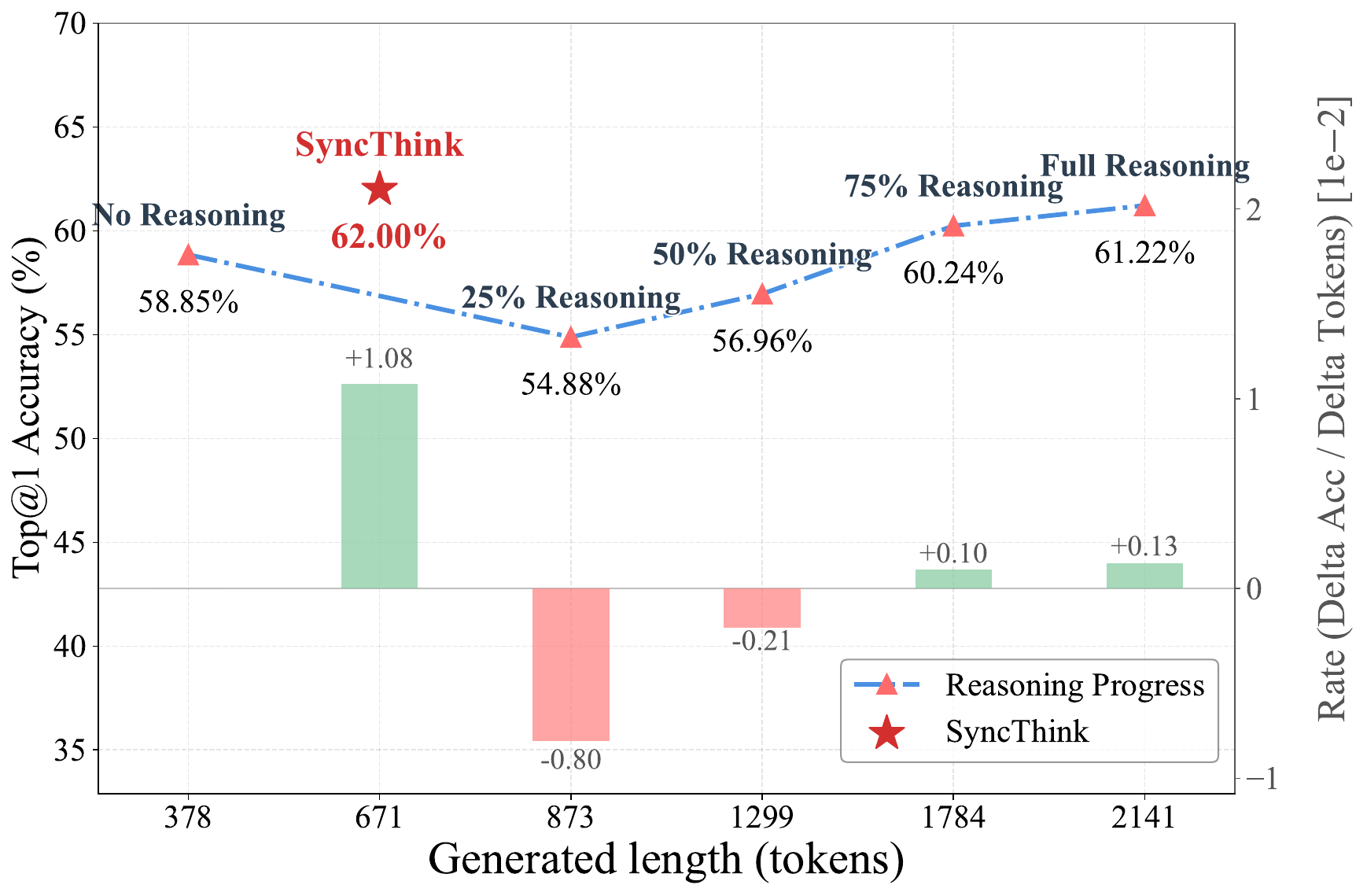}
  \caption{\small
  \textbf{Efficiency-Accuracy Trade-off.} The \textcolor{blue}{blue curve} depicts the base model's capability boundary. \textbf{SyncThink} (\textcolor{red}{red star}) significantly surpasses this limit, occupying the \textbf{top-left Pareto frontier} with higher accuracy and reduced token usage. The bottom bars represent the \textbf{marginal reasoning efficiency} ($\Delta \text{Acc}/\Delta \text{Tokens}$). Our method achieves the highest score (1.08), demonstrating that it effectively retains critical reasoning while pruning redundancy. More results across four datasets can be viewed in Appendix \ref{app:pareto_frontier_analysis}.}
  \label{fig:illustration}
\end{figure}

Large language models (LLMs)~\cite{vaswani2017attention, zhang2025will} achieve strong reasoning via Chain-of-Thought (CoT) prompting~\cite{wei2022chain}, but long CoT traces substantially increase inference cost. 
As Figure~\ref{fig:illustration} shows, accuracy gains quickly saturate beyond a token budget threshold, yielding diminishing returns. 
Consistent with prior findings that models over-think~\cite{chen2025think}, often reach correct answers early~\cite{liu2025answer}, and can perform well with concise drafts~\cite{xu2025chain}, this points to a mismatch between generated length and true reasoning demand, leaving many tokens redundant for the final prediction.

However, existing solutions remain limited. Training-based methods~\cite{yao2023react, bi2024forest, ye2024multi, zhang2024self} require additional fine-tuning or specialized data, reducing deployability. Training-free heuristics such as Answer Convergence~\cite{liu2025answer} and sampling-based stopping rules~\cite{sui2025stop} rely on black-box consistency checks and repeated probing, often adding extra inference cost without leveraging the model's intrinsic signals.

We formalize this inefficiency as \textbf{Cognitive Lag}: the model has already reached a sufficient reasoning state, yet continues generating redundant tokens. From a \textbf{Universal Reasoning Bottleneck} view, attention/saliency analyses (\cref{fig:attention_maps,fig:attention_maps}) show that a transition token (e.g., \texttt{</think>}) acts as an \textbf{information bridge} that compresses the reasoning history into a decidable state. We therefore monitor reasoning by tracking the step-wise logit rank of this token (\cref{fig:main_graph}). Macro results on BBH further reveal that accuracy saturates in an ``optimal truncation zone'' (starting at $\sim$60\% of the reasoning progress), well before the model emits its intrinsic termination signal.

To bridge this lag, we propose \textbf{SyncThink}, a training-free decoding framework motivated by the \textbf{Universal Reasoning Bottleneck}. 
We view the shift from reasoning to answering as a semantic boundary, which is explicitly marked in reasoning-specialized models (e.g., \texttt{</think>} in DeepSeek-R1). 
SyncThink treats this transition token as a high-fidelity probe: by tracking its logit \textit{rank} and \textit{entropy}, we trigger \emph{dynamic early stopping} when the model reaches information saturation.

Experiments show that SyncThink achieves a better efficiency--accuracy trade-off. 
In Figure~\ref{fig:illustration}, SyncThink reaches 62.00\% Top@1 accuracy with $\sim$656 tokens, outperforming full CoT (61.22\%, $\sim$2141 tokens) while reducing the token budget by 69.4\% ($3.21\times$ speedup). 
We further quantify this gain with a \textbf{reasoning efficiency rate} (bottom bars), defined as the accuracy improvement per additional token over the non-reasoning baseline ($\Delta \text{Acc}/\Delta \text{Tokens}$). 
Fixed truncation yields negligible or negative returns and full reasoning shows diminishing marginal utility (0.13), whereas SyncThink achieves the highest efficiency score (1.08), indicating it preserves high-value reasoning while removing redundant ``over-thinking'' tokens.

    


Our contributions are as follows:
\begin{itemize}[leftmargin=*, noitemsep, topsep=2pt]
    \item We define \textbf{Cognitive Lag} in CoT reasoning and show, via micro/macro analyses, that accuracy saturates well before the model’s termination signal.
    \item We introduce a \textbf{Universal Reasoning Bottleneck} view and, supported by attention/saliency evidence (\cref{fig:main_graph,fig:attention_maps,fig:attention_maps}), establish \texttt{</think>} as a high-fidelity probe of the reasoning-to-answer transition.
    \item We propose \textbf{SyncThink}, a training-free decoding method that monitors intrinsic logit signals to dynamically truncate redundant reasoning, improving efficiency while preserving or improving accuracy across tasks.
\end{itemize}

\section{Related Work}

\subsection{LLMs Reasoning}
Chain-of-Thought (CoT) prompting~\cite{wei2022chain} fundamentally enhances LLMs by decomposing complex tasks into intermediate reasoning steps. This paradigm has been further extended by strategies like Self-Consistency~\cite{wang2022self}, which improves robustness via path aggregation, and Tree-of-Thoughts~\cite{yao2023tree}, which introduces structured planning. Most recently, advanced models such as OpenAI’s o1, Alibaba’s QwQ~\cite{qwq32b}, and DeepSeek’s R1~\cite{guo2025deepseek} integrate reflective capabilities—including trial-and-error and self-correction—directly into the reasoning process, moving beyond static prompting to dynamic inference~\cite{shinn2023reflexion}.


\subsection{CoT Compression}
%
Efficient CoT generation is primarily addressed through two paradigms: training-based compression and inference-time optimization~\cite{qu2025survey,sui2025stop}. Training approaches aim to minimize reasoning overhead by either pruning explicit tokens based on difficulty~\cite{han2024token,zhang2025lightthinker} or mapping thoughts to latent states~\cite{chen2024not,qu2025optimizing}. 
In contrast, inference-time methods like Sketch-of-Thought~\cite{aytes2025sketch,xu2025chain} achieve efficiency by generating concise reasoning drafts, bypassing the need for model retraining.

Our SyncThink aligns with prior work on inference-time CoT optimization, but differs in that it is entirely training-free and model-agnostic, making it readily applicable to DeepSeek-R1 distilled models. Instead of learning to compress reasoning, SyncThink dynamically identifies the reasoning state from the model’s logits and lightweightly truncates redundant reasoning once completion signal is detected, while a logit-driven completion mechanism restores output quality and formatting with minimal overhead.

\begin{figure*}[t]

    \centering
    
    \begin{subfigure}{\textwidth}
        \centering
        \includegraphics[width=\linewidth]{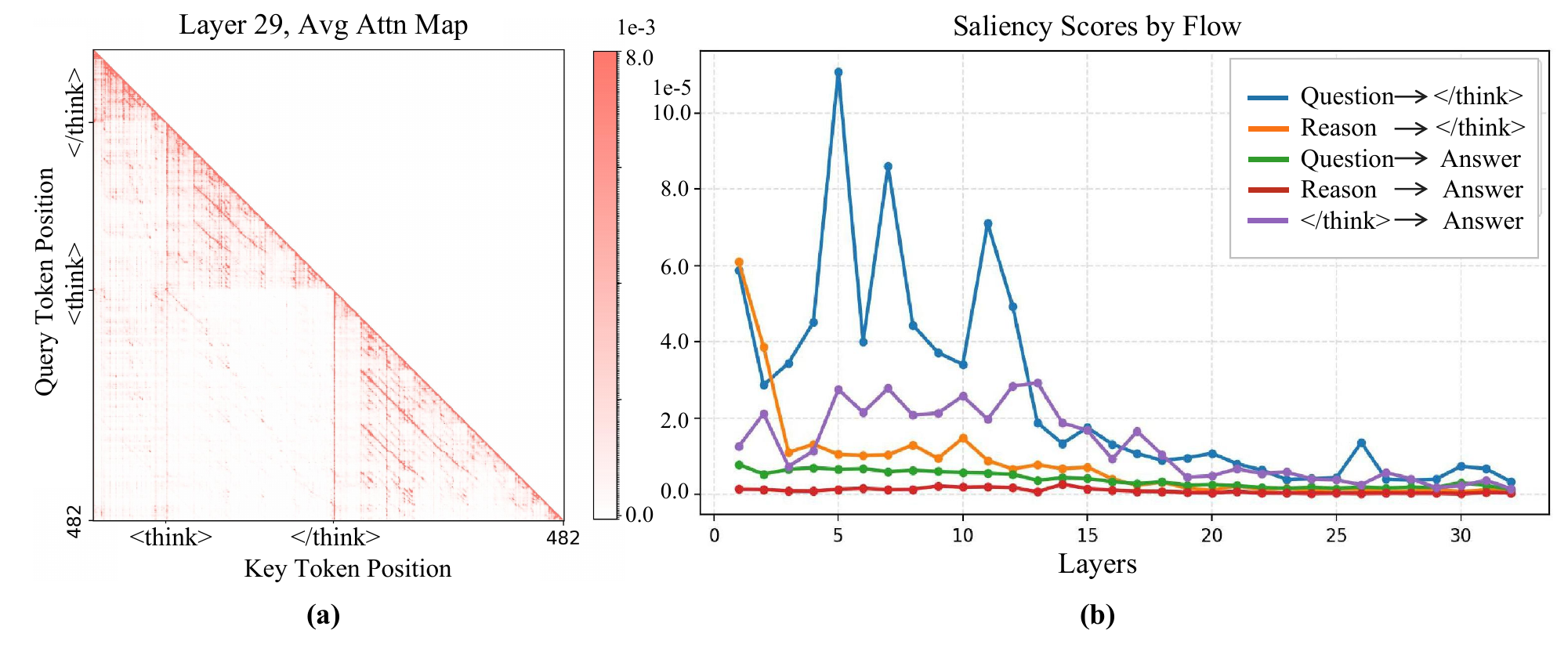}

    \end{subfigure}
    
   \caption{Empirical evidence of the \textbf{Universal Reasoning Bottleneck} on a GSM8K example. 
    (a) At Layer 29, attention concentrates on the \texttt{</think>} boundary rather than reasoning tokens (more layers in Fig.~\ref{fig:attention_maps_appendix}). 
    (b) Saliency analysis shows that information transfer from Reasoning to Answer is negligible (\textcolor{red}{red line}), while the dominant pathway is mediated by \texttt{</think>} (\textcolor{mplPurple}{purple line}), suggesting reasoning is compressed into this transition token.
    }
    \label{fig:attention_maps}
\end{figure*}

\section{Methodology}

We propose \textbf{SyncThink}, a \textit{training-free} framework designed to optimize CoT reasoning efficiency by eliminating redundant computation. To ground this approach in the model's intrinsic mechanics, our methodology proceeds through a rigorous analytical pipeline:

\begin{enumerate}[leftmargin=*, label=(\arabic*), nosep]
    \item \textbf{Mechanism Identification (\S\ref{subsec:analysis}):} Through attention and saliency analysis, we first validate the \texttt{</think>} token as a critical \textbf{Information Bottleneck}, establishing it as a high-fidelity probe for monitoring internal reasoning states.
    \item \textbf{Phenomenon Quantification (\S\ref{sec:rank_dynamics}):} By tracking the temporal dynamics of this probe, we reveal the \textbf{Cognitive Lag}. Our micro-level phase analysis and macro-level accuracy statistics jointly demonstrate that answer saturation precedes the model's explicit termination.
    \item \textbf{Algorithm Design (\S\ref{subsec:solution}):} Building on these insights, we formulate SyncThink, a heuristic strategy that leverages dynamic rank and entropy thresholds to intervene precisely at the point of information saturation.
\end{enumerate}

\subsection{Problem Formulation: The Efficiency-Accuracy Trade-off}
\label{subsec:formulation}

Given a question $q$, an LLM generates a reasoning chain $r = (r_1, \dots, r_L)$ followed by an answer $a$. While $r$ generally enhances the conditional probability $P(a|q, r)$, the computational cost scales linearly with $L$. 
Recent observations in the literature~\cite{chen2025think, liu2025answer} indicate that this performance gain exhibits \textit{diminishing returns}: accuracy tends to saturate rapidly after an initial rise, implying that extending $L$ beyond a critical point yields negligible gains. 

Therefore, our objective is to find an optimal truncation point $\hat{t}$ that maximizes accuracy while minimizing cost:
\begin{equation}
\hat{t}=\arg\max_{t}\left(A(t)-\alpha\cdot \mathrm{Cost}(t)\right)
\end{equation}

Since the true accuracy $A(t)$ is unobservable during inference, we seek an intrinsic proxy from the model's internal state to estimate reasoning sufficiency.

\begin{figure*}[t]
    \centering
    \includegraphics[width=0.90\linewidth]{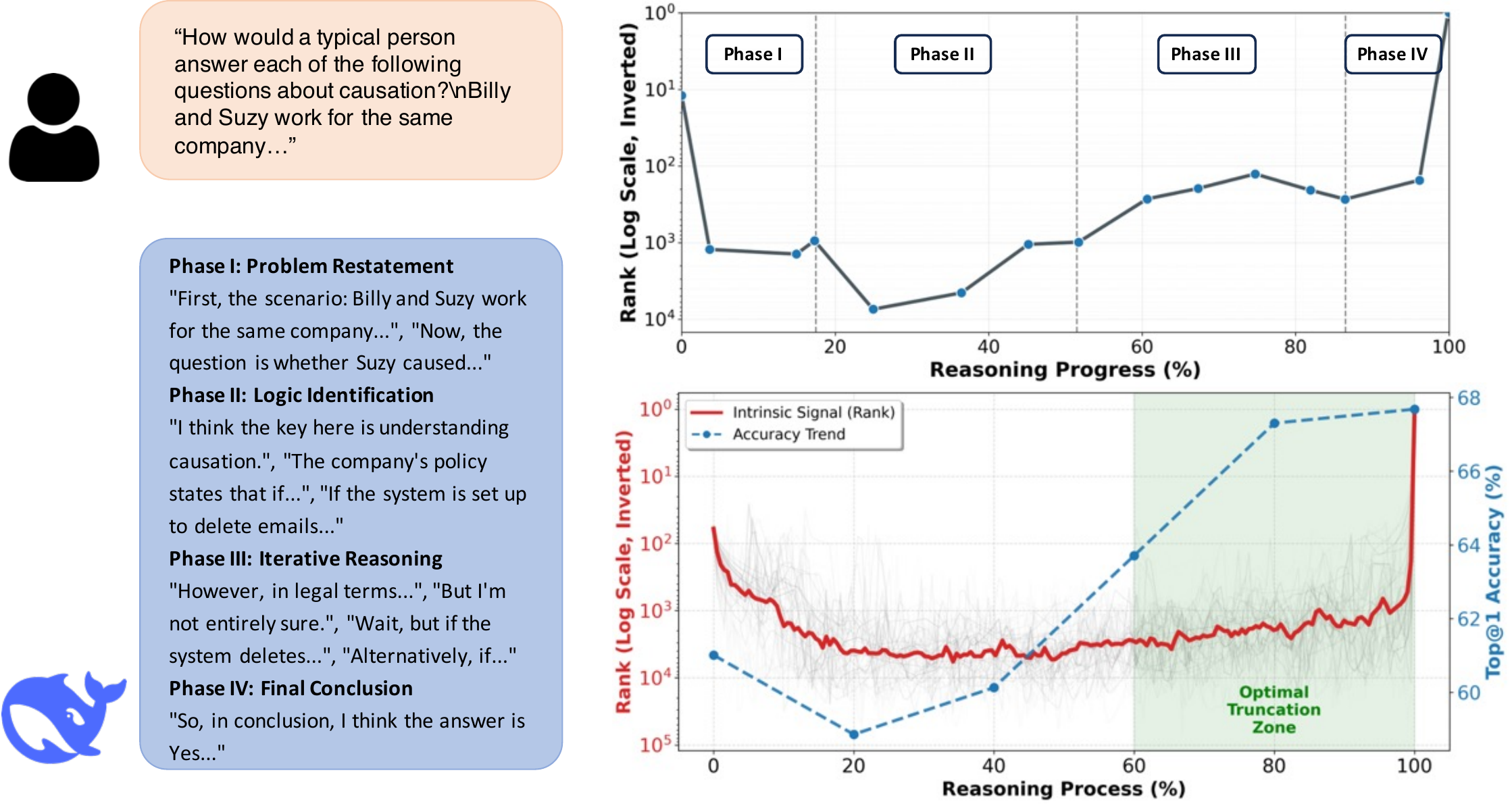}
    \caption{\small
    \textbf{The rationale behind SyncThink.} 
    (Top) \textbf{Micro-Dynamics:} The \texttt{</think>} rank provides a real-time signal that tracks the four reasoning phases in \S\ref{sec:rank_dynamics}. 
    (Bottom) \textbf{Macro-Statistics:} On BBH, we observe a \emph{cognitive lag}: truncation accuracy (\textcolor{blue}{blue}) saturates within the shaded \emph{optimal truncation zone} well before the model’s intrinsic termination signal (\textcolor{red}{red}).}
    \label{fig:main_graph}
\end{figure*}

\subsection{Attention Analysis: The Information Bottleneck}
\label{subsec:analysis}

To identify an intrinsic proxy, we first examine internal attention maps, as self-attention governs information flow in Transformers. 
As shown in Figure~\ref{fig:attention_maps}, deep-layer attention allocates negligible mass to individual reasoning tokens and instead concentrates on the \texttt{</think>} boundary. 
This observation suggests that (i) the reasoning span contains substantial redundancy, and (ii) \texttt{</think>} serves as an \textbf{information bottleneck} where critical context is compressed and accumulated for answer generation.

Further, to validate these hypotheses, we employ Saliency Score analysis following \citet{wang2023label}. This framework quantifies the contribution of specific attention regions by measuring the sensitivity of the loss function to masked perturbations via a first-order Taylor approximation:
\begin{equation}
    \Delta L \approx \alpha \left\langle A_{h,l} \odot \frac{\partial L}{\partial A_{h,l}}, \, M \right\rangle_F .
\end{equation}

    

    

This approximation reveals that saliency is essentially the element-wise product of the attention activations and their gradients within the masked region. 

As illustrated in Figure~\ref{fig:attention_maps}, this analysis yields three key insights regarding the reasoning mechanism:

\begin{itemize}[leftmargin=*, noitemsep, topsep=2pt]
    \item \textbf{The \texttt{</think>} Token as an Information Bridge:} The information flow follows a ``Reasoning $\to$ \texttt{</think>} $\to$ Answer'' path. Direct saliency from reasoning tokens to the answer is negligible, confirming that \texttt{</think>} serves as a critical mediator that aggregates context before generation.
    
    \item \textbf{Early Aggregation and Logic Construction:} Saliency peaks in shallow layers (1--2) for context compression, while core logical decision-making is centralized in the middle layers (5--15). This indicates that the primary information processing occurs in the early-to-mid stages of the network.
    
    \item \textbf{Completion of Information Integration:} In deeper layers ($>20$), saliency scores across all key paths diminish significantly (quiescence). This confirms that critical information migration to \texttt{</think>} is fully concluded, rendering further explicit retrieval of reasoning history redundant.
\end{itemize}

Our saliency analysis identifies </think> as a critical information bottleneck where reasoning content is distilled, while intermediate tokens exhibit rapid saliency decay. This phenomenon is not limited to specific instances and is analyzed in detail in Appendix \ref{app:attention_sink}. Consequently, the model's primary reliance on this distilled signal supports our strategy of using it for dynamic early stopping.

\subsection{The Temporal Dynamics of Reasoning Completeness}\label{sec:rank_dynamics}
Building on the identification of \textit{where} information concentrates, we now examine the temporal dynamics to determine \textit{when} reasoning becomes sufficient. Specifically, we track the evolution of the \texttt{</think>} token's rank, defined as the rank of its logit among the full vocabulary at each decoding step. We use this step-wise logit rank as a lightweight proxy for the model’s internal reasoning state. As shown in Figure~\ref{fig:main_graph} (Top), the rank trajectory maps to four distinct cognitive phases:

\paragraph{Phase I: Problem Restatement (Sharp Descent).} 
The rank initially precipitates from the start to $>10^4$. This marks the transition from instruction encoding to reasoning initiation, indicating the model's commitment to generating a long-form chain rather than an immediate answer.

\paragraph{Phase II: Logic Identification (Initial Recovery).} 
A steady recovery to $\sim10^3$ follows, corresponding to the \textit{Orientation} stage. As the model structures the problem space and identifies a viable logical path, the uncertainty regarding termination decreases.

\paragraph{Phase III: Iterative Reasoning (Volatile Plateau).} 
The rank oscillates between $10^2$ and $10^3$. These fluctuations reflect \textit{Self-Reflection}: local peaks suggest the completion of sub-steps, while subsequent drops indicate the activation of verification or backtracking mechanisms during conflict resolution.

\paragraph{Phase IV: Final Conclusion (Linear Ascent).} 
Upon resolving the internal debate, the trajectory shifts to a decisive, linear ascent to Top-1. Unlike the fluctuations in Phase III, this irreversible rise serves as a deterministic signal that the reasoning process is concluded.

This analysis highlights the limitation of fixed token budgets, which fail to distinguish between necessary formulation (Phase II) and potentially redundant exploration (Phase III). Our framework leverages these rank trends to detect latent termination signals, enabling dynamic truncation during periods of "overthinking" to prevent redundant computation.

\paragraph{Macro-Statistics: The Cognitive Lag.}
We further aggregate reasoning trajectories across the BBH dataset to analyze macro-level trends (Figure~\ref{fig:main_graph}, Bottom). Here, the red curve depicts the median rank of the \texttt{</think>} token overlaid on individual sample traces (grey), while the blue dashed line indicates the Top-1 accuracy achievable by truncating generation at the corresponding progress percentage.

This macro-view reveals a critical decoupling between model capability and generation policy, termed \textbf{"Cognitive Lag"}. 
Specifically, as reasoning progresses beyond 60\%, the truncation accuracy (blue) enters a saturation plateau, recovering to near-optimal levels. 
Paradoxically, the model's intrinsic termination signal (red) remains deeply suppressed (rank $\sim10^3$) throughout this phase, indicating a lack of confidence to terminate despite having effectively solved the problem. 
This misalignment drives the model to engage in prolonged, low-utility reasoning to pursue negligible performance gains. 

Consequently, a dynamic intervention mechanism is required to decouple termination from the model's inertia, intervening precisely when information saturation is detected.

\subsection{SyncThink: A Dynamic Reasoning Termination Method Based on the Logit\label{subsec:solution}}

Our investigation reveals a mechanism of \textbf{Progressive Information Condensation}, where the model aggregates reasoning logic into evolving hidden states and employs the \texttt{</think>} token as a ``summary embedding'' of completeness. This dynamic reflects a \textbf{Mechanism Competition} between \textit{Exploration} and \textit{Conclusion} modes, where a rising rank of \texttt{</think>}—indicative of state saturation—signals the model's readiness to transition. Addressing the noise sensitivity caused by the magnitude dominance of raw logits, we demonstrate that the token \textbf{Rank} serves as an \textit{intrinsic proxy for information sufficiency}. 

Motivated by this, we introduce a heuristic dynamic truncation mechanism. When the rank of the terminator token ascends to a sufficiently high position, we manually inject \texttt{</think>} to terminate the reasoning process. We formulate this strategy as follows:

\begin{equation}
\label{eq:decision}
\mathbb{I}_{\text{stop}}(t) = 
\begin{cases} 
1 & \text{if } \mathcal{R}_t(\texttt{</think>}) \le \tau(t, \mathbf{p}_t) \\
0 & \text{otherwise}
\end{cases}
\end{equation}

Here, $\mathcal{R}_t(\texttt{</think>})$ denotes the rank of the terminator token within the current probability distribution $\mathbf{p}_t$ (0-indexed). The dynamic threshold $\tau(t, \mathbf{p}_t)$ is modulated by both temporal pacing and model uncertainty:

\begin{equation}
\label{eq:threshold}
\tau(t, \mathbf{p}_t) = \lfloor \beta(t) \cdot \exp(-\lambda \cdot \mathcal{H}(\mathbf{p}_t)) \rfloor
\end{equation}

The components of the threshold are defined as follows:

\begin{itemize}[leftmargin=*,itemsep=2pt, topsep=0pt, parsep=0pt]
    \item \textbf{Time-Dependent Pacing $\beta(t)$:} We directly employ $t$ as the time-scale control factor. As $t$ increases, the threshold rises accordingly, aligning with the intuition that longer reasoning chains are increasingly prone to redundancy. Additionally, we impose an upper bound $T_{\text{max}}$ to prevent the threshold from becoming excessively loose in extended sequences, thereby avoiding indiscriminate truncation.
    \item \textbf{Entropy-Aware Scaling $\mathcal{H}(\mathbf{p}_t)$:} We compute the Shannon entropy of the next-token distribution to adjust for model confidence:
    \begin{equation}
        \mathcal{H}(\mathbf{p}_t) = - \sum_{v \in V} p_t(v) \log p_t(v)
    \end{equation}
    where $V$ is the vocabulary. High entropy (uncertainty) reduces the threshold via the term $e^{-\lambda \mathcal{H}}$, tightening the constraint to prevent premature truncation during exploration phases~\cite{wang2025beyond}. Conversely, low entropy encourages commitment to a conclusion.
\end{itemize}

Once the condition $\mathbb{I}_{\text{stop}}(t) = 1$ is met, we manually inject the \texttt{</think>} token and proceed to the answer generation phase.

\section{Experimental Setup}

\paragraph{Datasets}\label{sec:datasets}
To comprehensively evaluate our proposed method across diverse reasoning and knowledge-intensive scenarios, we conduct experiments on the following four benchmark datasets:
\begin{itemize}[leftmargin=*,itemsep=2pt, topsep=0pt, parsep=0pt]
\item \textbf{GSM8K}~\cite{cobbe2021training}: Evaluates multi-step arithmetic reasoning via grade-school math problems.

\item \textbf{MMLU}~\cite{hendrycks2020measuring}: Assesses general knowledge across 57 diverse domains, ranging from elementary to professional levels.

\item \textbf{GPQA}~\cite{rein2024gpqa}: Tests high-level conceptual understanding with graduate-level scientific questions.

\item \textbf{BBH}~\cite{suzgun2022challenging}: Stress-tests models on hard-to-solve tasks involving symbolic and logical planning.

\end{itemize}

%
\paragraph{Backbone.}
We select models in the 7B--14B range, reflecting practical constraints while maintaining strong reasoning capabilities.

\textit{Qwen2.5-7B/14B}~\cite{yang2024qwen2}: Leading open-source models with robust reasoning performance. 
\textit{LLaMA3.1-8B}~\cite{grattafiori2024llama}: A balanced model optimized for efficient instruction following. 
Identical decoding settings are applied across all baselines for consistent comparison.

\paragraph{Baselines.}
We compare SyncThink against five baselines:
(1) \textbf{Full CoT}: Generates complete reasoning traces, serving as the inference-time quality upper bound;
(2) \textbf{No CoT}: Directly generates answers without reasoning;
(3) \textbf{Fixed-Ratio Truncation}: A heuristic that stops reasoning at a predetermined length ratio;
(4) \textbf{Answer Convergence}~\cite{liu2025answer}: A probing method that triggers early stopping when intermediate answer predictions stabilize across $k$ consecutive segments.
Additionally, on GSM8K, we include an \textbf{SFT Oracle/Upper Bound} (fine-tuned on gold reasoning traces) to benchmark the maximum achievable performance via supervised learning.

All baselines are evaluated with identical prompts and answer extraction, and we match the generation budget across methods whenever applicable,
enabling an apples-to-apples comparison of accuracy and efficiency.

\paragraph{Metrics.}
We evaluate each method from both effectiveness and efficiency perspectives using three primary metrics:
Top-1 accuracy (Top@$1$$\uparrow$), wall-clock inference time (Time$\downarrow$), and the number of generated tokens (Tokens$\downarrow$).
Top@$1$ accuracy is computed as the exact-match rate under a unified answer parser for each benchmark.

\paragraph{Decoding parameters.}
Unless otherwise specified, all results are obtained under \textbf{greedy decoding} (temperature $=0$ and no stochastic sampling)
with \texttt{max\_new\_tokens}$=8192$ for all methods.
Under this setting, generation is (near-)deterministic given the same model, tokenizer, prompts, and inputs; consequently, we report results
from a \textbf{single deterministic run} per configuration. More specific details can be seen in Appendix~\ref{app:inference_details}.
\textbf{For our truncation strategy} in Eq.~\ref{eq:threshold}, we tune $\lambda$ on the held-out validation set and fix it to $\lambda=0.8$, which is then applied unchanged to all test benchmarks; we also set $\beta(t)=t$ (the decoding step) to gradually relax the early-exit constraint as generation proceeds.

\input{tables/dataset_table_int}

\begin{figure*}
    \centering
    \includegraphics[width=1.0\linewidth]{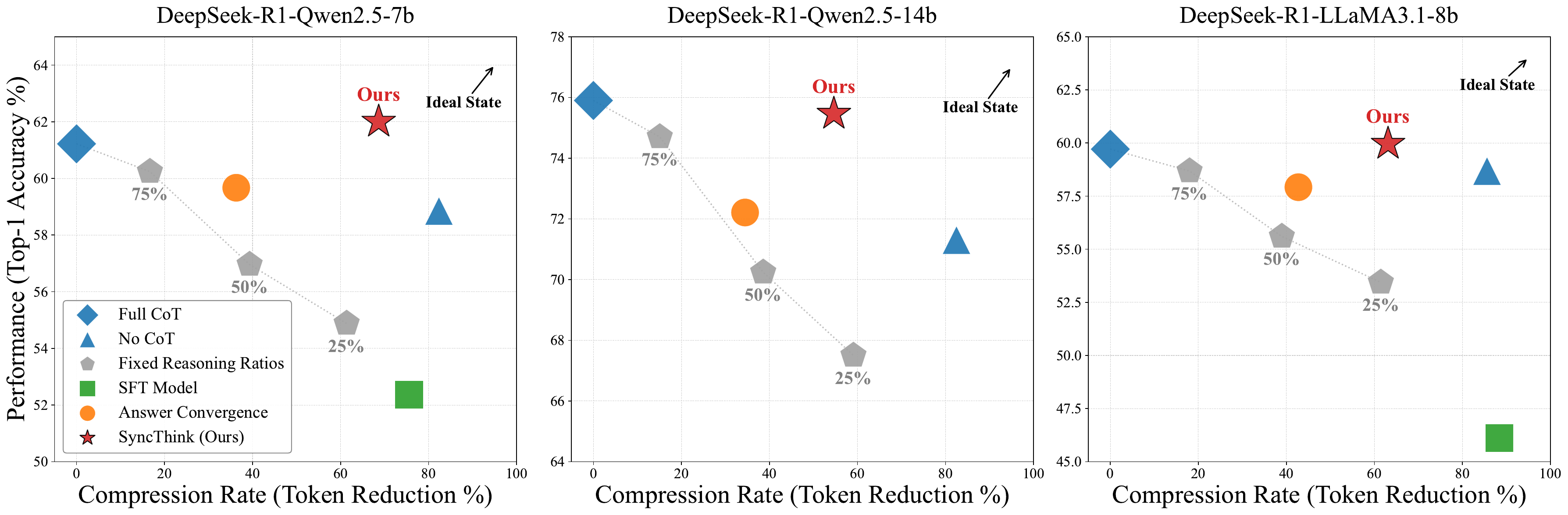}
    \caption{\textbf{Efficiency-Accuracy Trade-off.} SyncThink demonstrates superior Pareto efficiency compared to Full CoT, SFT, and static truncation baselines. The red star indicates our method achieves the best balance between token reduction and model performance across all evaluated models.}
    \label{fig:pareto_frontier}
\end{figure*}

\section{Results and Analysis}

\subsection{Main Results}
%

Table~\ref{tab:main_results} presents the performance comparison across four benchmarks. As indicated by the average results, \textbf{SyncThink} achieves a superior trade-off between reasoning quality and inference efficiency. Notably, \textbf{SyncThink} yields the highest average Top@1 accuracy (\textbf{62.00\%}), surpassing the \textbf{Full Reasoning} baseline (61.22\%) while requiring only $\sim$30\% of the token generation cost (655.80 vs. 2141.25 tokens). This suggests that our dynamic truncation mechanism effectively identifies the critical reasoning path without redundant computation.

\subsection{Mitigating Over-Thinking via Information Synchronization}
\label{sec:analysis_syncthink}

Models often succumb to ``over-thinking,'' where valid reasoning degrades into hallucination due to delayed termination. \textbf{SyncThink} addresses this by synchronizing generation with internal information saturation.

\paragraph{Case Study: Pruning Harmful Redundancy.}
Figure~\ref{fig:case_study_gpqa} presents a representative GPQA example where \emph{more} reasoning becomes \emph{worse}. 
The Full CoT baseline reaches a correct intermediate state and is already poised to answer. However, instead of committing, it continues generating additional derivations and self-corrections (e.g., \textit{``Wait, but I need to relate this to...''}), gradually drifting into an unnecessary re-formulation of the problem and ultimately outputting an incorrect option (C). 
In contrast, \textbf{SyncThink} monitors the logit dynamics of the reasoning-state transition token \texttt{</think>} and identifies the onset of this detrimental redundancy. 
As highlighted in the figure, SyncThink triggers an early exit at the \emph{earlier} \texttt{</think>} boundary (upper marker), where the model's information has already saturated, and directly produces the correct answer (A), avoiding the later over-thinking branch that leads to error.

\paragraph{Adaptive Truncation.}
Our method also adapts to task complexity. On simple benchmarks (GSM8K), \textbf{SyncThink} applies conservative truncation, preserving linear reasoning paths. Conversely, on complex tasks (GPQA), it exhibits a higher truncation rate. This confirms that \textbf{SyncThink} selectively targets the ``over-thinking'' loops prevalent in difficult reasoning, rather than indiscriminately shortening all responses.

\subsection{Universality of the Reasoning Bottleneck}
\label{sec:universal_bottleneck}

A natural question is whether our reliance on the \texttt{</think>} signal is specific to DeepSeek-R1-style training. 
We argue that it reflects a more general \emph{mechanistic} pattern in LLMs: attention heads often concentrate probability mass on a small set of ``sink'' tokens that stabilize internal state and aggregate long-range context. 
In CoT decoding, the shift from a divergent \emph{reasoning phase} to a convergent \emph{answering phase} requires such a semantic boundary, which effectively compresses the preceding rationale into a decidable state. 
DeepSeek-R1 makes this boundary explicit via \texttt{</think>}, while other models may realize the same function implicitly through discourse markers (e.g., \textit{Therefore}, \textit{Thus}) or role separators. 
Therefore, SyncThink leverages \texttt{</think>} as a concrete instance of a \textbf{Reasoning Attention Sink}~\cite{xiao2023efficient}, and the core idea remains applicable whenever an equivalent transition token can be identified.
We provide a more detailed discussion in Appendix~\ref{app:attention_sink}.

\subsection{Efficiency and Pareto Trade-off}
\textbf{SyncThink} effectively shifts the accuracy--efficiency Pareto frontier outward. On average, it matches \textbf{Full Reasoning} accuracy (62.00\% vs. 61.22\%) while reducing token consumption by $\sim$70\% (655.80 vs. 2141.25). This contrasts with static heuristics that merely trade accuracy for speed, and \textbf{SFT baselines} which fail due to surface-level mimicry—often resulting in either aggressive over-truncation or degenerate infinite loops. The benefits are most pronounced on complex tasks like GPQA, where \textbf{SyncThink} (38.38\%) outperforms both Full Reasoning (30.30\%) and No Reasoning (32.82\%) by mitigating error accumulation. Overall, \textbf{SyncThink} offers a superior operating point, translating token savings into practical latency and memory reductions.

\paragraph{Architectural Robustness.}
Table~\ref{tab:main_results} confirms \textbf{SyncThink}'s consistency across diverse architectures (LLaMA and Qwen families, 7B--14B). 
Notably, it achieves a $\sim$3$\times$ speedup on Qwen-7B with preserved accuracy. 
This validates the \texttt{</think>} logit gap as a robust, model-agnostic indicator for reasoning completeness.

\begin{figure}[htbp]
    \centering
    \includegraphics[width=0.5\textwidth]{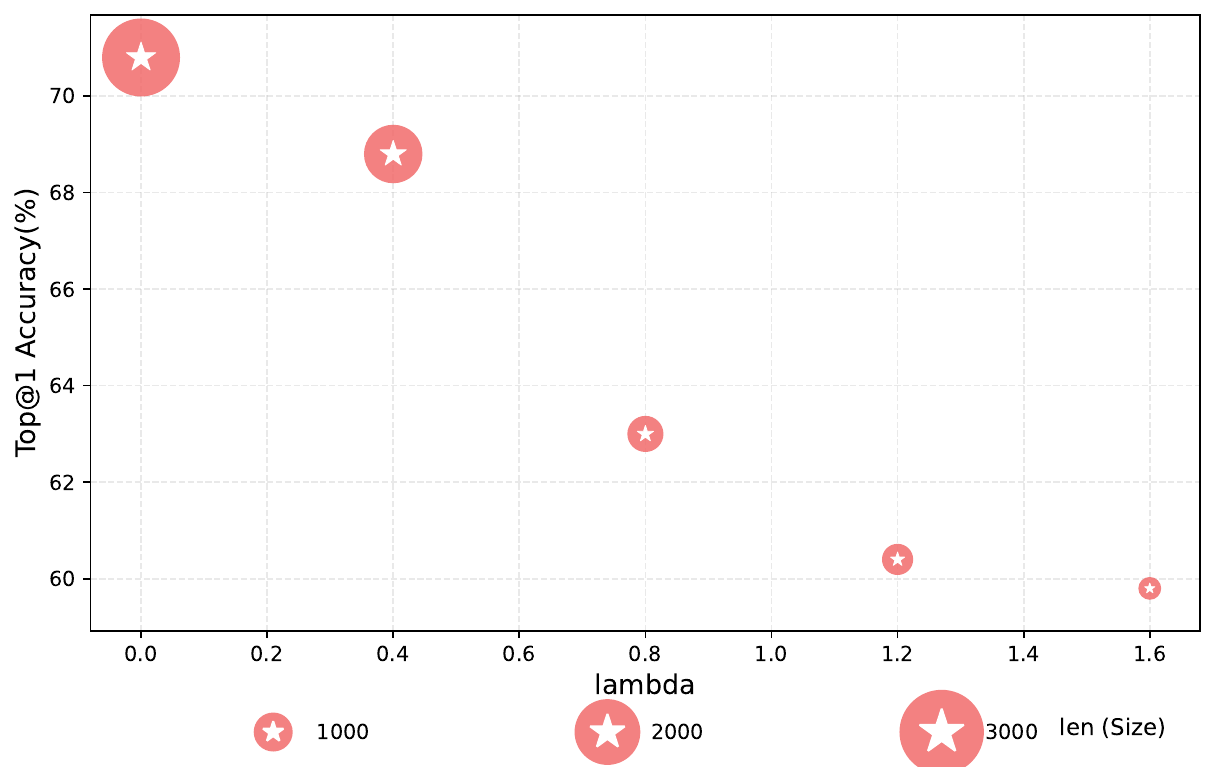}
    \caption{Ablation study on $\lambda$.} 
    \label{fig:ablation_figure}  
\end{figure}


\subsection{Parameter Analysis}
\label{sec:param_lambda}

To avoid overfitting to evaluation benchmarks, we tune the entropy weight $\lambda$ on a held-out \textsc{MATH500}~\cite{lightman2023let} validation set, sweeping $\lambda$ and measuring Top@1 accuracy and average generation length (Figure~\ref{fig:ablation_figure}).

Increasing $\lambda$ monotonically shortens generations, confirming it controls truncation aggressiveness. Accuracy drops as $\lambda$ grows, yielding a clear accuracy--length Pareto trade-off. Thus, $\lambda$ provides a simple knob to select the desired efficiency--performance operating point.



\subsection{Performance Across Diverse Tasks} \textbf{SyncThink} generalizes effectively across diverse domains, surpassing the \textbf{Full Reasoning} baseline with over 60\% computational savings. Notably, this is achieved using a unified hyperparameter configuration without task-specific fine-tuning. 
On complex benchmarks like GPQA, our method successfully mitigates ``over-thinking'' by halting generation before hallucinations occur: for Qwen2.5-7B, SyncThink improves accuracy from 30.30\% (Full Reasoning) to 38.38\%. 
This stability confirms that SyncThink captures intrinsic reasoning patterns. \textbf{The hyperparameters thus serve not as sensitive tuning variables, but as a consistent lever for users to prioritize either extended reasoning for marginal gains or aggressive truncation for efficiency}.

\section{Conclusion}
We propose \textbf{SyncThink}, a training-free and model-agnostic decoding framework that synchronizes CoT generation with the model’s intrinsic reasoning sufficiency signals. 
Our key insight is that these signals already indicate \emph{where} and \emph{when} reasoning becomes sufficient. 
Attention analyses reveal an \textbf{information bottleneck} at the reasoning boundary token (e.g., \texttt{</think>}), while temporal dynamics show the model can \emph{sense} its reasoning state. 
Due to training-induced generation bias, the emitted termination is often delayed, causing \textbf{Cognitive Lag}---the model is already decidable yet continues producing redundant tokens. 
SyncThink addresses this misalignment with a lightweight decision rule that aligns decoding with information saturation, reducing unnecessary computation while preserving answer quality.
Empirically, SyncThink matches \textbf{Full Reasoning} accuracy on average (62.00\% vs.\ 61.22\%) with $\sim$70\% fewer tokens (655.80 vs.\ 2141.25), and improves GPQA to 38.38\% (vs.\ 30.30\% Full; 32.82\% No).

%

\section{Limitations}
\label{sec:limitations}

While our proposed method demonstrates effectiveness, we acknowledge two aspects regarding the scope of our current analysis:

\begin{itemize}[leftmargin=*,itemsep=2pt, topsep=0pt, parsep=0pt]
    \item \textbf{Cross-Model Validation.}
    Our experiments mainly focus on models that expose an explicit reasoning-to-answer boundary (e.g., \texttt{</think>}). 
    As discussed in Appendix~\ref{app:attention_sink}, we view \texttt{</think>} as a surface realization of a more general attention-dynamics phenomenon (e.g., sink-like attractor behavior) around phase transitions, rather than a model-specific artifact.
    Due to time constraints, we do not conduct a large-scale cross-model study in this submission; comprehensive validation across diverse model families.

    \item \textbf{Scale Verification:} Our current experiments have not yet extended to \textbf{larger-scale models} (e.g., 70B+ parameters). As model behavior and attention redundancy often evolve with scale, further empirical study is needed to confirm the consistency of the information bottleneck phenomenon and the optimal configuration of our method in massive-scale settings.
\end{itemize}

\section{Ethical Considerations}

We use publicly available datasets and model checkpoints under licenses permitting research use; license terms and restrictions are summarized in Section~\ref{sec:datasets}. 
All artifacts are used in accordance with their intended purposes specified by the original providers. 
We also use an LLM only for writing assistance and language polishing.


\clearpage

\appendix

\clearpage

\section{Implementation Details}
\label{app:inference_details}

\subsection{Inference Configuration}
We conduct all inference experiments using the \texttt{transformers} library on a high-performance cluster node equipped with \textbf{8 $\times$ NVIDIA H800 GPUs}. To ensure reproducibility, we set the random seed to $5$ across all runs. The specific hyperparameters for generation are listed below:

\begin{itemize}[leftmargin=*]
    \item \textbf{Decoding Strategy:} We utilize greedy decoding (\texttt{do\_sample=False}) to eliminate randomness in the generation trajectory and ensure deterministic evaluation results.
    \item \textbf{Precision:} All models are loaded in \texttt{bfloat16} precision to balance computational efficiency and numerical stability.
    \item \textbf{Batch Size:} We set the inference batch size to $1$ to simulate a real-world streaming latency scenario.
    \item \textbf{Max Generation Length:} The maximum number of new tokens is set to 8192  to accommodate the full reasoning chain of the baseline models.
\end{itemize}

\subsection{Latency Measurement Protocol}
To rigorously evaluate the efficiency of our proposed \textbf{SyncThink} method, we define the inference latency metric to include the full end-to-end processing time. Specifically, the reported time $T_{total}$ is calculated as:
\begin{equation}
    T_{total} = T_{gen} + T_{metric} + T_{eval}
\end{equation}
where:
\begin{itemize}
    \item $T_{gen}$ denotes the raw model generation time.
    \item $T_{metric}$ represents the computational overhead introduced by our method, including the real-time extraction and calculation of token logits and attention entropy.
    \item $T_{eval}$ includes the time required for the final answer extraction and correctness verification.
\end{itemize}
This comprehensive measurement ensures that the reported speedup ($3.21\times$) reflects the net performance gain after accounting for all computational costs associated with our dynamic monitoring mechanism.

\section{Pareto Frontier Analysis of SyncThink}
\label{app:pareto_frontier_analysis}

\begin{figure*}[!t]

    \centering
    
    \begin{subfigure}{0.45\textwidth}
        \centering
        \includegraphics[width=\linewidth]{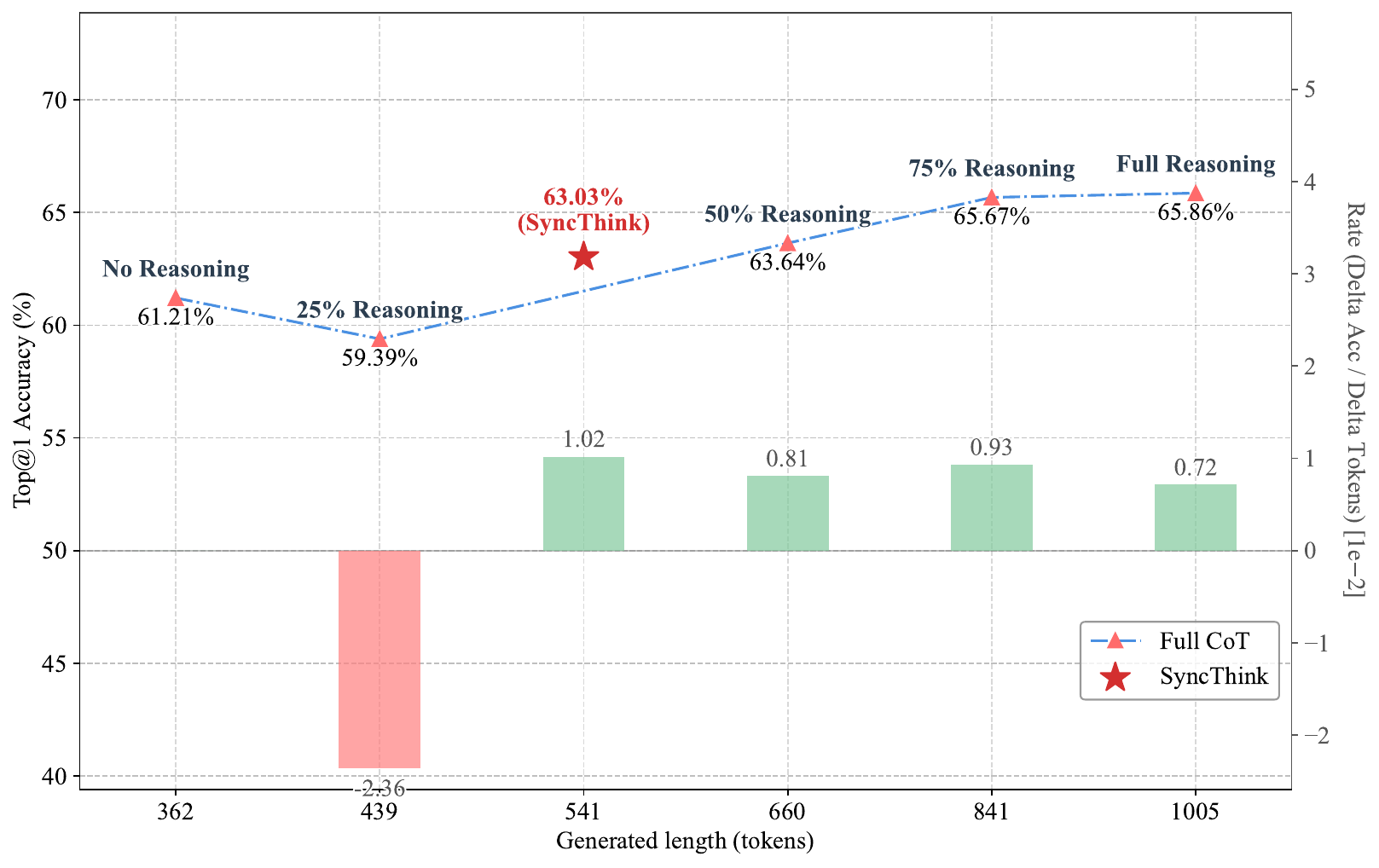}
        \caption{BBH}
    \end{subfigure}
    \hfill
    \begin{subfigure}{0.45\textwidth}
        \centering
        \includegraphics[width=\linewidth]{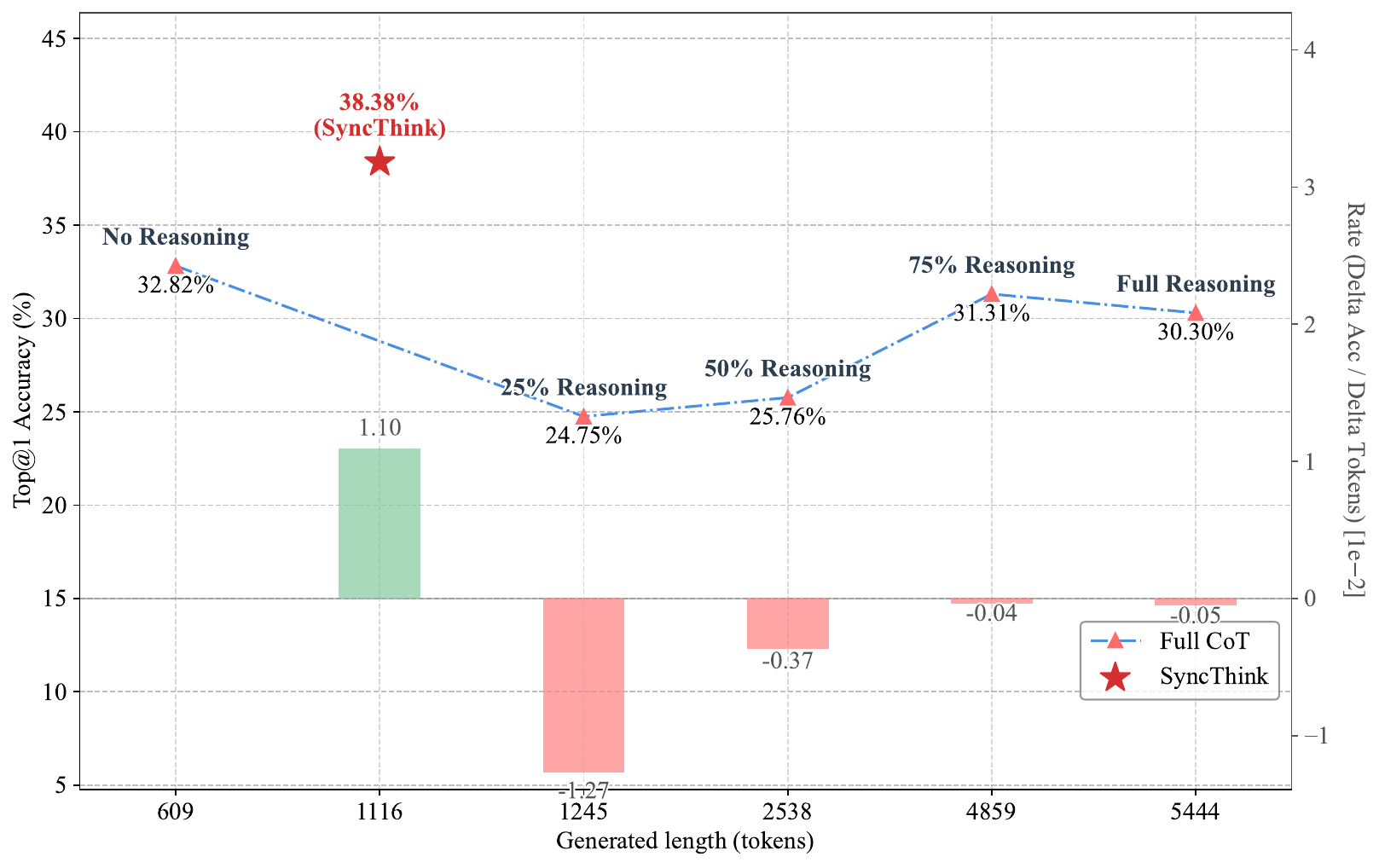}
        \caption{GPQA}
    \end{subfigure}
    \hfill
    \begin{subfigure}{0.45\textwidth}
        \centering
        \includegraphics[width=\linewidth]{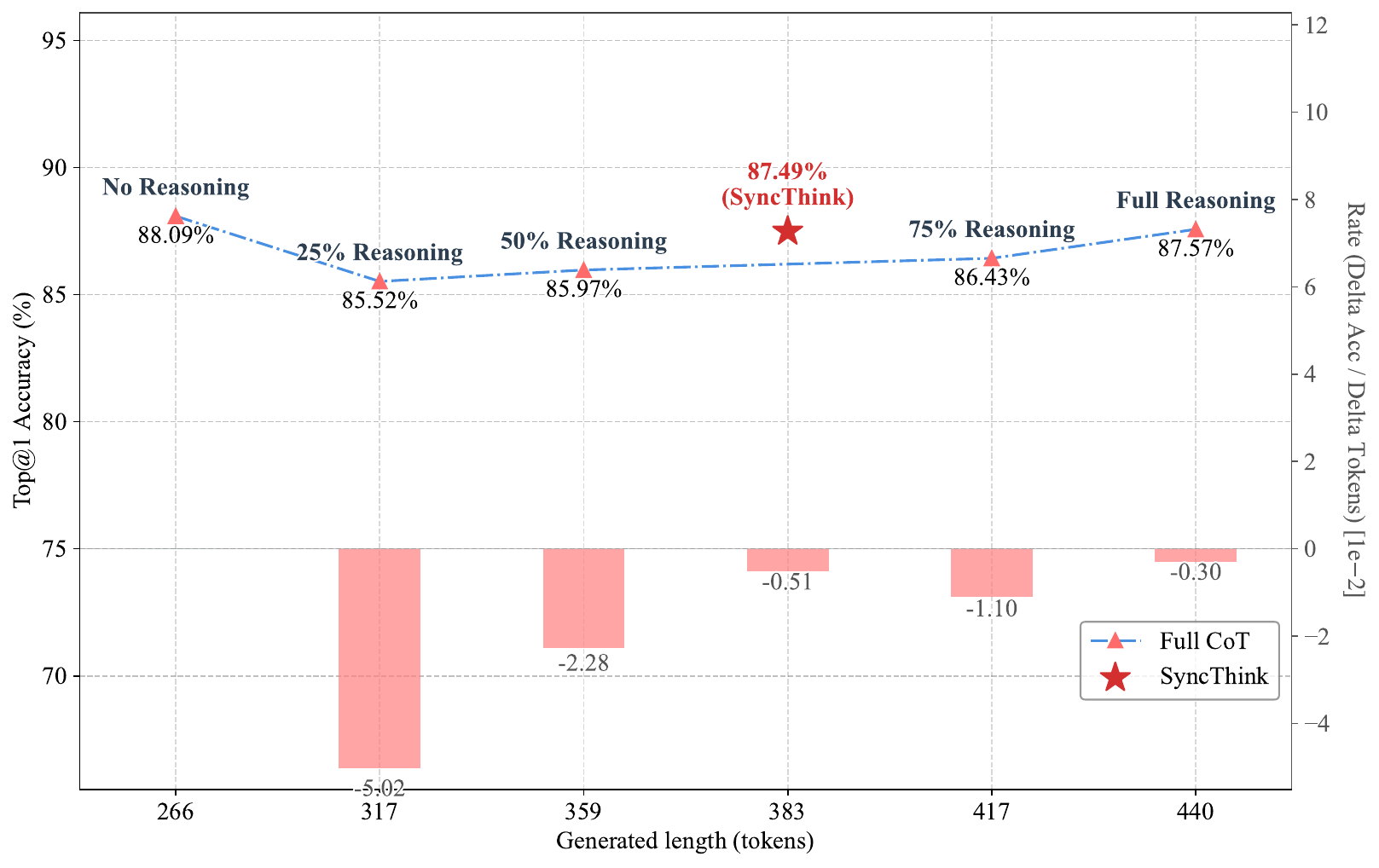}
        \caption{GSM8K}
    \end{subfigure}
    \hfill
    \begin{subfigure}{0.45\textwidth}
        \centering
        \includegraphics[width=\linewidth]{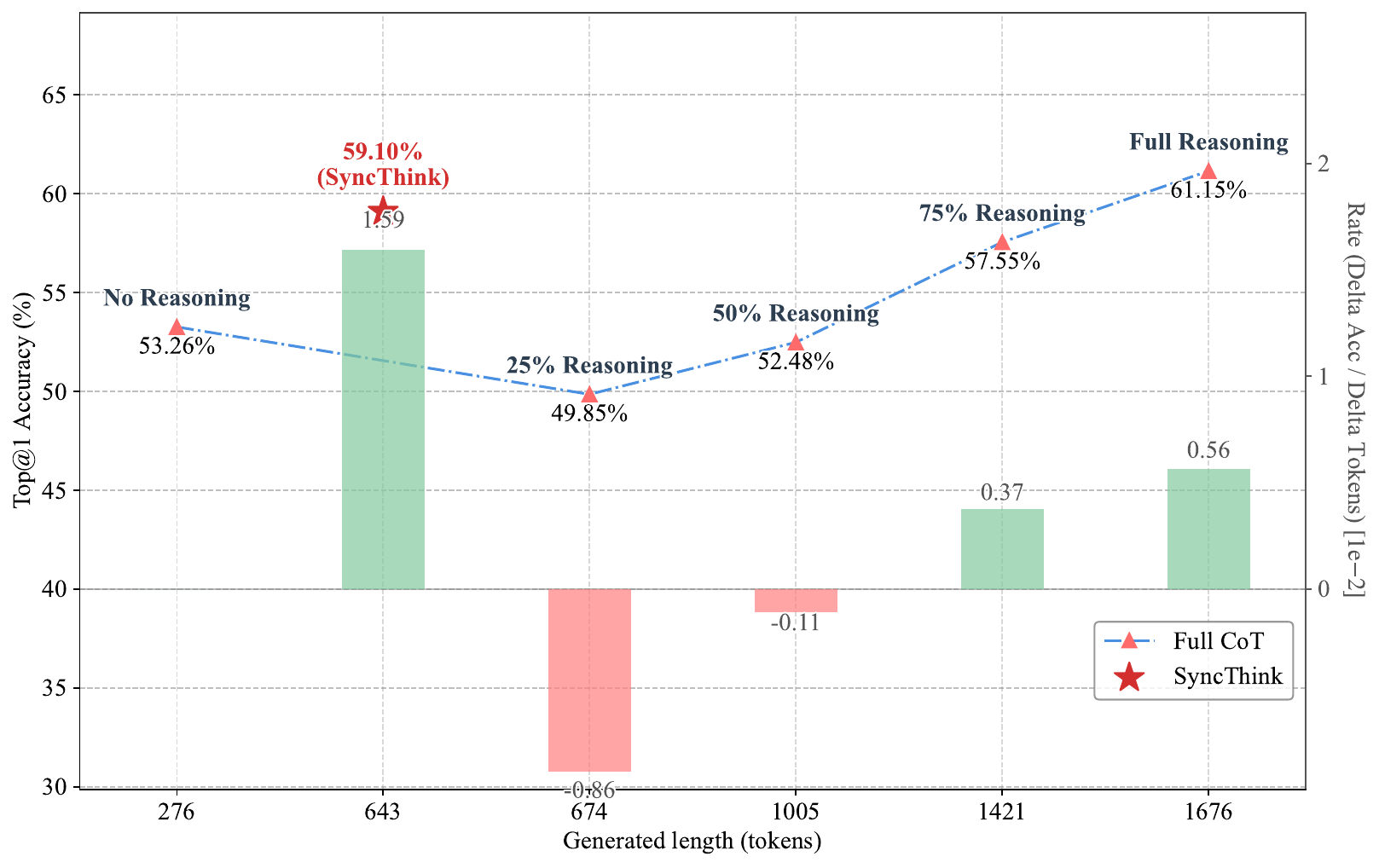}
        \caption{MMLU}
    \end{subfigure}

    \caption{\textbf{Accuracy--length trade-off of SyncThink on DeepSeek-R1-Distill-Qwen-7B.}
    Blue curves denote fixed-ratio truncation baselines on full CoT traces, while the red star marks \textbf{SyncThink}. Across BBH, GPQA, GSM8K, and MMLU, \textbf{SyncThink} consistently lies in the upper-left region, achieving higher accuracy with fewer generated tokens, with the largest gain on GPQA.}
    \label{fig:pareto_dataset_analysis}
\end{figure*}

To examine the accuracy--efficiency trade-off, we sweep the truncation ratio of a fixed-ratio baseline and plot accuracy against the truncation ratio on four benchmarks.
This sweep induces a set of candidate operating points, whose upper envelope forms an \emph{empirical Pareto frontier} for uniform truncation policies.
As shown in Figure~\ref{fig:pareto_dataset_analysis}, \textbf{SyncThink} consistently lies in the upper-left region relative to this frontier across all datasets, indicating a strictly better operating point: it achieves higher accuracy under the same truncation ratio, or equivalently reaches comparable accuracy with fewer reasoning tokens.
Importantly, this gain cannot be replicated by simply selecting a different global truncation ratio.
Instead, it stems from \textbf{SyncThink}'s instance-adaptive, logit-driven termination rule, which detects when additional reasoning becomes redundant and reallocates the reasoning budget more effectively than any single fixed-ratio strategy.

\section{The Universality of the Reasoning Bottleneck}
\label{app:attention_sink}

In this section, we discuss the theoretical grounding of our method, arguing that the effectiveness of the \texttt{</think>} token is not an artifact of specific model training (e.g., DeepSeek-R1) but a manifestation of a universal mechanism in Large Language Models (LLMs).

\paragraph{The Attention Sink Hypothesis.}
Recent studies on efficient streaming LLMs have proposed the \textit{Attention Sink} hypothesis~\cite{xiao2023efficient}, which suggests that attention heads tend to allocate massive attention scores to specific ``sink'' tokens (often initial tokens or special separators) to maintain internal state stability and aggregate historical context.
In the specific context of Chain-of-Thought (CoT) reasoning, the transition from a \textit{reasoning phase} to an \textit{answering phase} necessitates a semantic boundary---a token or sequence that compresses the preceding divergent rationale into a converged state for answer generation.

\paragraph{Function over Form.}
While DeepSeek-R1 explicates this boundary as \texttt{</think>}, we argue that other models employ implicit equivalents to fulfill the same functional role.
For instance, standard base models often rely on linguistic markers such as ``\textit{Therefore}'', ``\textit{So}'', or ``\textit{Thus}'' to signal this state shift. similarly, instruction-tuned models may utilize special tokens like \texttt{<|start\_header\_id|>} or specific role separators.
Consequently, the \texttt{</think>} token in our \textbf{SyncThink} framework essentially functions as a specialized \textit{Reasoning Attention Sink}.
It signals the phase shift from a high-entropy \textit{exploration state} to a low-entropy \textit{determination state}.

This theoretical perspective implies that \textbf{SyncThink} is generalizable to a broader range of CoT-capable models.
The core methodology---detecting the saturation of reasoning information via a bottleneck token---remains valid, provided one identifies the model-specific transition token (the ``sink''), which distinguishes itself only in surface form rather than functional utility.


\begin{figure*}[t]
    \includegraphics[width=1.0\linewidth]{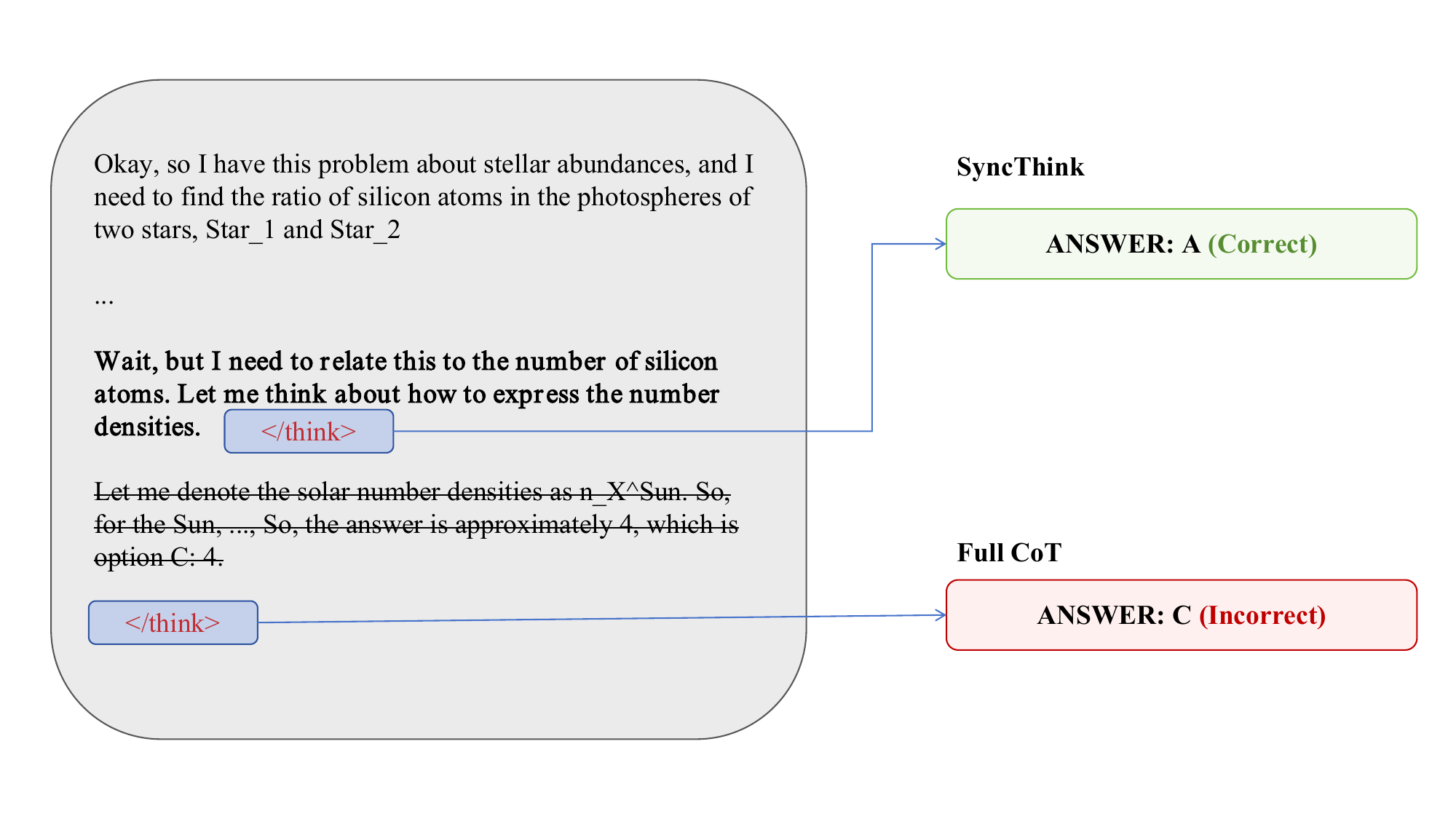}
    \caption{Case study on preventing model over-thinking. While the Full CoT path drifts into erroneous reasoning after a correct intermediate state, our SyncThink method detects the onset of this detrimental redundancy. By truncating the generation early, our method successfully avoids the subsequent error and retains the correct answer.}
    \label{fig:case_study_gpqa}
\end{figure*}

\clearpage
\begin{figure*}[!t]

    \centering
    
    \begin{subfigure}{0.35\textwidth}
        \centering
        \includegraphics[width=\linewidth]{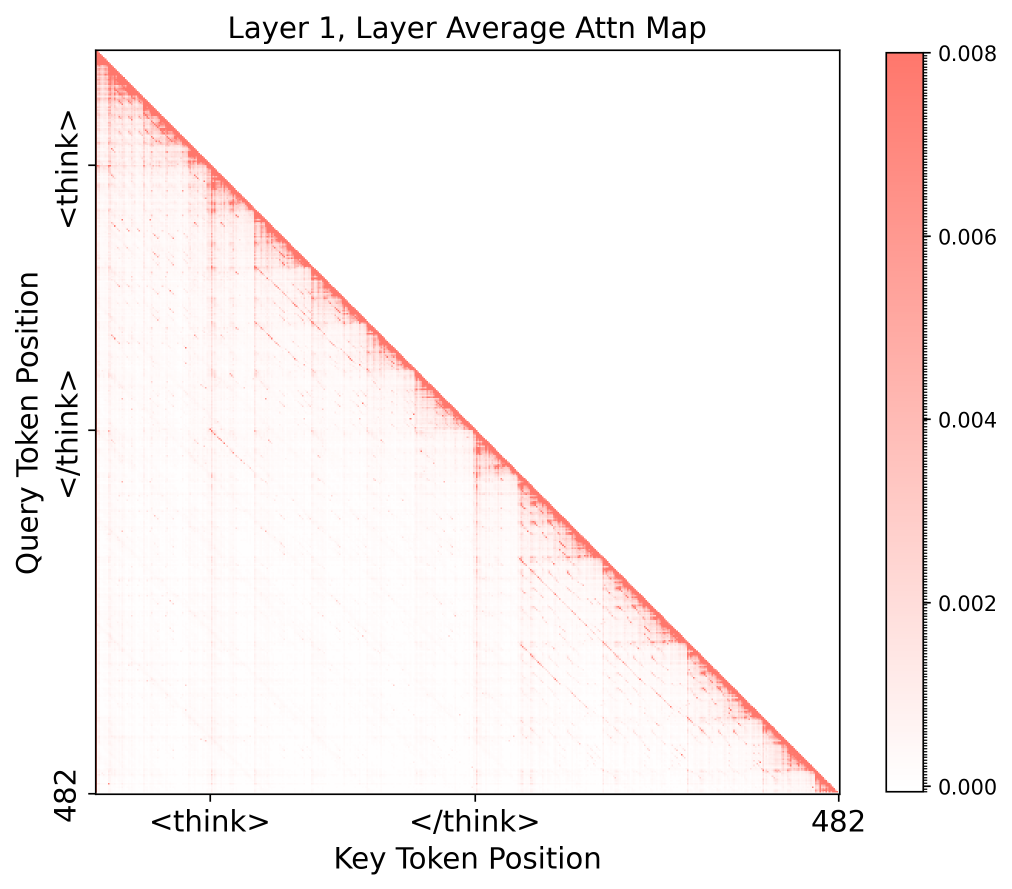}
        \caption{Layer 1}
    \end{subfigure}
    \hfill
    \begin{subfigure}{0.35\textwidth}
        \centering
        \includegraphics[width=\linewidth]{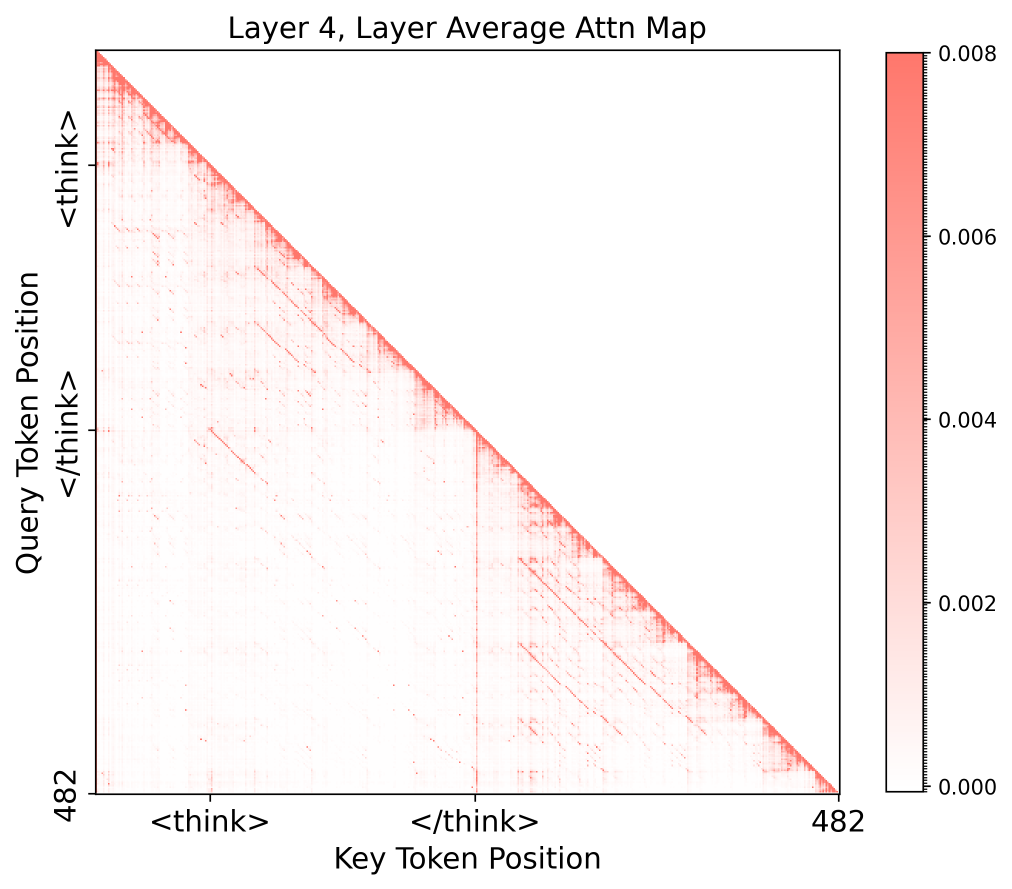}
        \caption{Layer 4}
    \end{subfigure}
    \hfill
    \begin{subfigure}{0.35\textwidth}
        \centering
        \includegraphics[width=\linewidth]{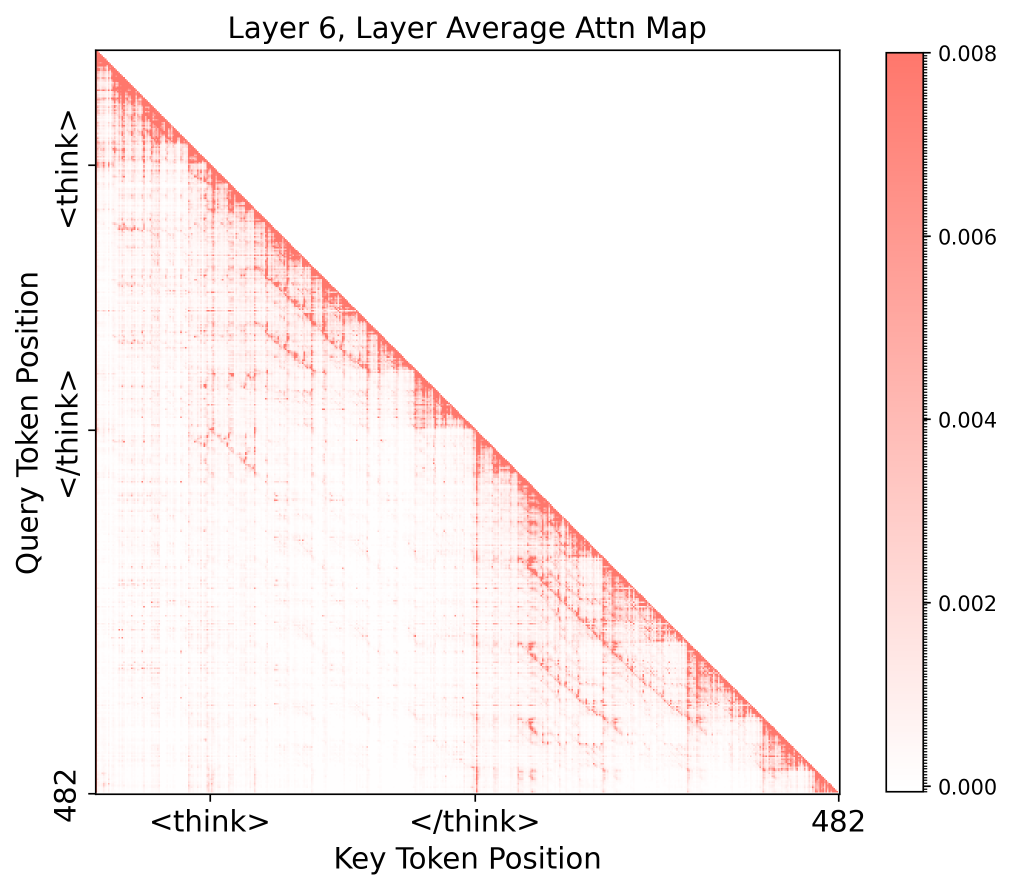}
        \caption{Layer 6}
    \end{subfigure}
    \hfill
    \begin{subfigure}{0.35\textwidth}
        \centering
        \includegraphics[width=\linewidth]{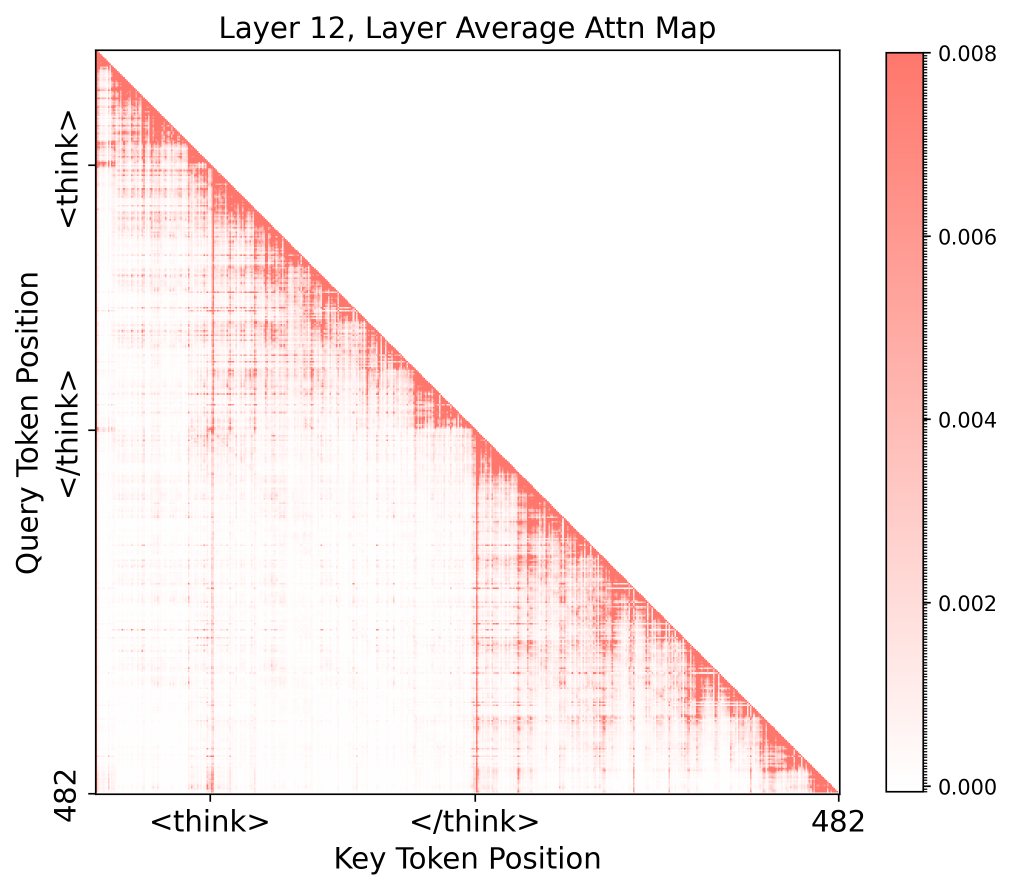}
        \caption{Layer 12}
    \end{subfigure}
    
    \vspace{0.5em}
    
    \begin{subfigure}{0.35\textwidth}
        \centering
        \includegraphics[width=\linewidth]{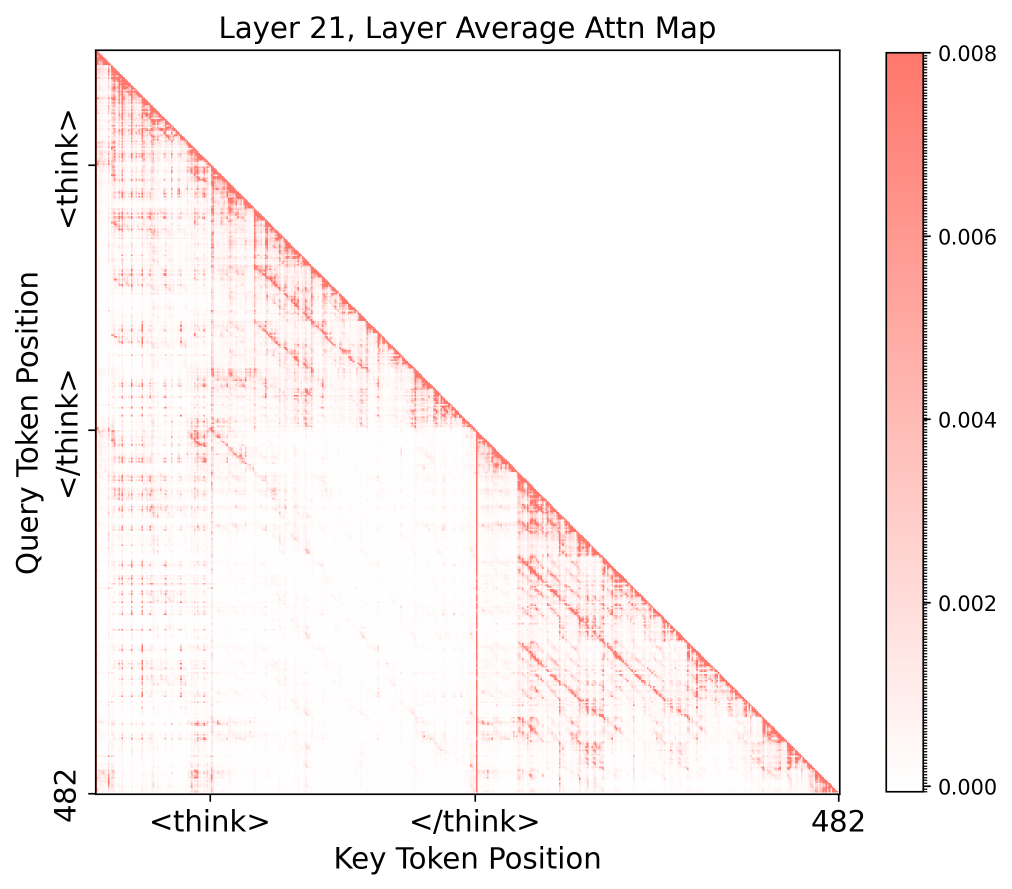}
        \caption{Layer 21}
    \end{subfigure}
    \hfill
    \begin{subfigure}{0.35\textwidth}
        \centering
        \includegraphics[width=\linewidth]{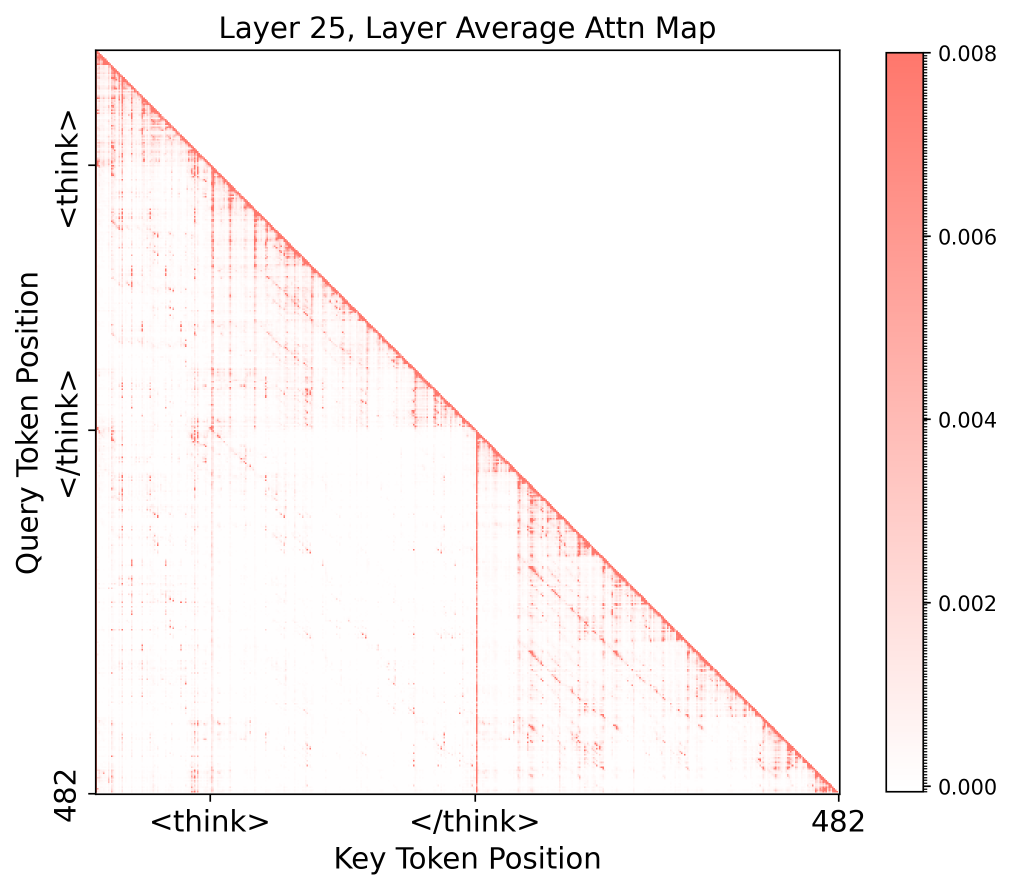}
        \caption{Layer 25}
    \end{subfigure}
    \hfill
    \begin{subfigure}{0.35\textwidth}
        \centering
        \includegraphics[width=\linewidth]{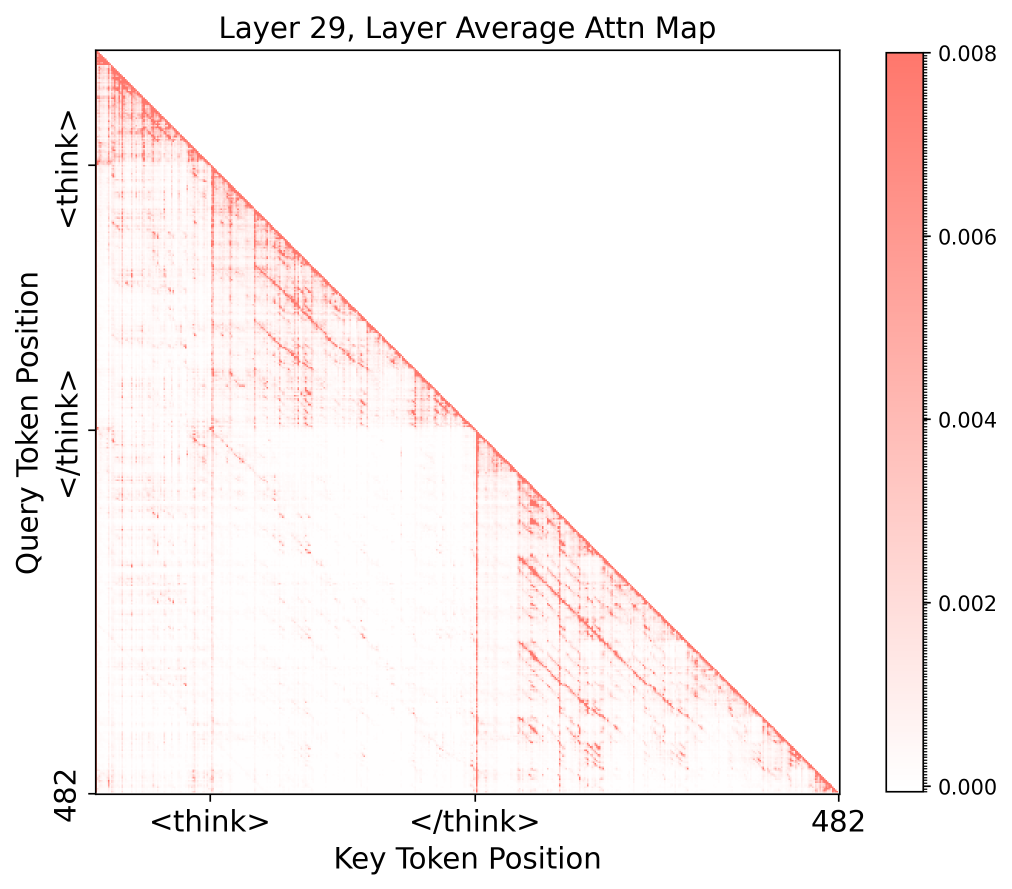}
        \caption{Layer 29}
    \end{subfigure}
    \hfill
    \begin{subfigure}{0.35\textwidth}
        \centering
        \includegraphics[width=\linewidth]{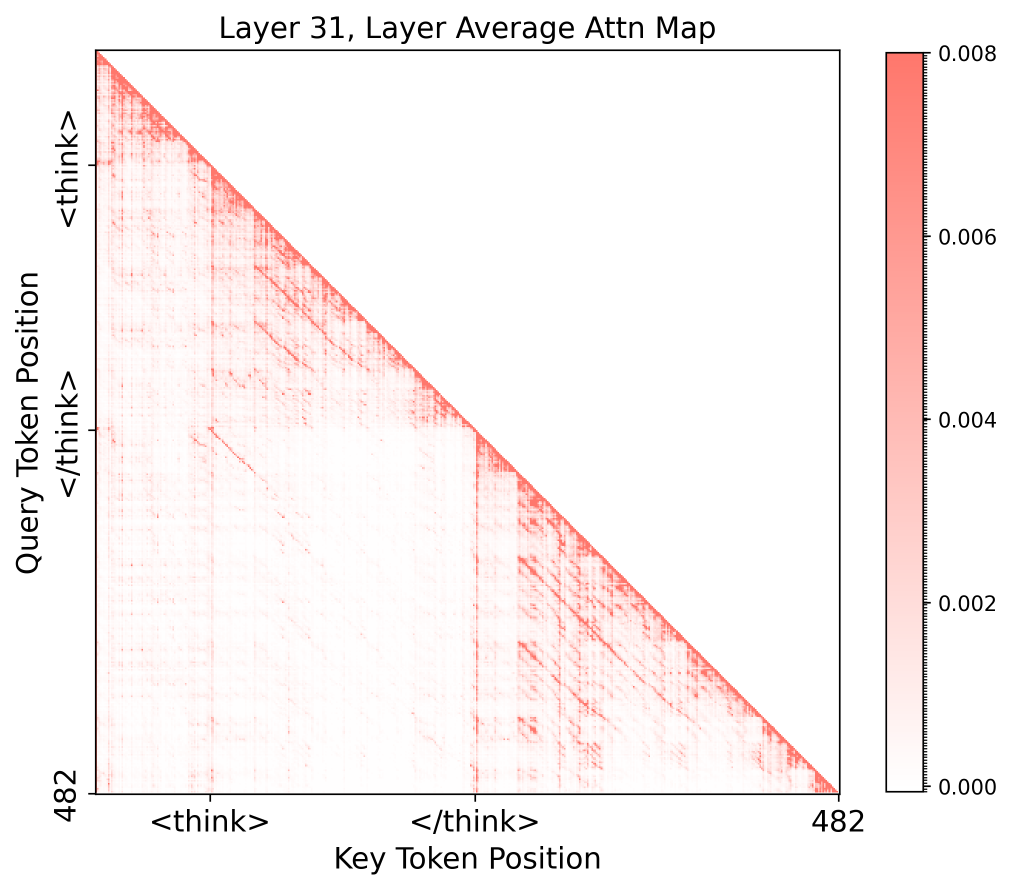}
        \caption{Layer 31}
    \end{subfigure}
    
    \caption{Attention heatmaps across different layers of DeepSeek-R1-Distill-LLaMA-8B on a GSM8K sample~\cite{cobbe2021training}. Tokens within the <$\mathrm{think}$>...<$\mathrm{/think}$> span receive uniform attention in early layers, but deeper layers gradually shift focus to the boundary tokens, indicating information migration and compression of reasoning content. Similar observations can be found in other models and datasets
    }
    \label{fig:attention_maps_appendix}
\end{figure*}

\end{document}

%% file: tables/dataset_table_int.tex
\definecolor{pale_green}{HTML}{ECF5E4}

\begin{table*}[t!]
\centering
\scriptsize
\setlength{\tabcolsep}{3.5pt} 
\resizebox{\textwidth}{!}{%
\begin{tabular}{l|ccc|ccc|ccc|ccc|ccc}
\toprule
\multirow{2}{*}{\textbf{Method}} & \multicolumn{3}{c|}{\textbf{GSM8K}} & \multicolumn{3}{c|}{\textbf{MMLU}} & \multicolumn{3}{c|}{\textbf{GPQA}} & \multicolumn{3}{c|}{\textbf{BBH}} & \multicolumn{3}{c}{\textbf{Avg Res}} \\
\cmidrule(lr){2-4} \cmidrule(lr){5-7} \cmidrule(lr){8-10} \cmidrule(lr){11-13} \cmidrule(lr){14-16}
 & \textbf{Top@1} & \textbf{Time} & \textbf{Tokens} & \textbf{Top@1} & \textbf{Time} & \textbf{Tokens} & \textbf{Top@1} & \textbf{Time} & \textbf{Tokens} & \textbf{Top@1} & \textbf{Time} & \textbf{Tokens} & \textbf{Top@1} & \textbf{Time} & \textbf{Tokens} \\
\midrule
\multicolumn{16}{c}{\textbf{DeepSeek-R1-Qwen2.5-7b}} \\
\midrule
Full Reasoning & 87.57 & 18.79 & 440 & 61.15 & 73.00 & 1676 & 30.30 & 231.58 & 5444 & 65.86 & 44.65 & 1005 & 61.22 & 92.01 & 2141 \\
No Reasoning & 88.09 & 11.39 & 266 & 53.26 & 12.01 & 276 & 32.82 & 25.90 & 609 & 61.21 & 16.08 & 362 & 58.85 & 16.35 & 378 \\
\cmidrule[0.3pt]{1-16}
Reasoning Ratio($r=0.25$) & 85.52 & 13.45 & 317 & 49.85 & 28.41 & 674 & 24.75 & 78.89 & 1879 & 59.39 & 19.51 & 439 & 54.88 & 35.07 & 827 \\
Reasoning Ratio($r=0.5$) & 85.97 & 15.22 & 359 & 52.48 & 42.61 & 1005 & 25.76 & 131.20 & 3170 & 63.64 & 29.35 & 660 & 56.96 & 54.60 & 1299 \\
Reasoning Ratio($r=0.75$) & 86.43 & 17.05 & 417 & 57.55 & 58.75 & 1421 & 31.31 & 184.43 & 4458 & 65.67 & 37.38 & 841 & 60.24 & 74.40 & 1784 \\
\cmidrule[0.3pt]{1-16}
SFT Model & 77.41 & 4.53 & 106 & 51.41 & 15.80 & 363 & 32.32 & 39.44 & 927 & 48.28 & 30.84 & 694 & 52.36 & 22.65 & 523 \\
Answer Convergence & 87.82 & 15.28 & 278 & 57.12 & 34.64 & 754 & 31.38 & 180.38 & 3919 & 62.36 & 27.21 & 503 & 59.67 & 64.38 & 1364 \\
\rowcolor{pale_green} \textbf{SyncThink} & 87.49 & 14.39 & 383 & 59.10 & 28.13 & 643 & \textbf{38.38} & \textbf{47.73} & \textbf{1116} & 63.03 & 24.46 & 541 & 62.00 & 28.68 & 671 \\
\midrule
\multicolumn{16}{c}{\textbf{DeepSeek-R1-Qwen2.5-14b}} \\
\midrule
Full Reasoning & 93.33 & 35.24 & 493 & 81.60 & 105.67 & 1465 & 43.43 & 378.28 & 5300 & 85.25 & 92.34 & 1287 & 75.90 & 152.88 & 2136 \\
No Reasoning & 92.04 & 18.07 & 251 & 74.88 & 21.65 & 299 & 40.91 & 49.92 & 698 & 77.37 & 17.66 & 247 & 71.30 & 26.83 & 374 \\
\cmidrule[0.3pt]{1-16}
Reasoning Ratio($r=0.25$) & 91.15 & 22.59 & 319 & 66.52 & 44.68 & 630 & 35.48 & 140.23 & 2021 & 76.88 & 37.40 & 525 & 67.50 & 61.23 & 874 \\
Reasoning Ratio($r=0.5$) & 91.62 & 27.09 & 379 & 70.03 & 63.90 & 888 & 36.92 & 217.19 & 3173 & 82.38 & 57.00 & 807 & 70.24 & 91.30 & 1312 \\
Reasoning Ratio($r=0.75$) & 92.12 & 31.17 & 459 & 76.80 & 87.14 & 1271 & 44.88 & 300.80 & 4455 & 85.00 & 73.78 & 1070 & 74.70 & 123.22 & 1814 \\
\cmidrule[0.3pt]{1-16}
SFT Model & 84.43 & 156.58 & 2191 & 64.23 & 329.49 & 4569 & 44.04 & 367.26 & 5145 & 70.59 & 341.63 & 4763 & 65.82 & 298.74 & 4167 \\
Answer Convergence & 92.23 & 33.49 & 381 & 75.33 & 70.29 & 744 & 41.91 & 188.38 & 3741 & 79.38 & 68.38 & 735 & 72.21 & 90.14 & 1400 \\
\rowcolor{pale_green} \textbf{SyncThink} & 93.03 & 31.59 & 442 & 77.99 & 61.59 & 852 & \textbf{46.97} & \textbf{126.29} & \textbf{1764} & 83.84 & 58.65 & 817 & 75.46 & 69.53 & 969 \\
\midrule
\multicolumn{16}{c}{\textbf{DeepSeek-R1-LLaMA3.1-8b}} \\
\midrule
Full Reasoning & 78.47 & 23.66 & 498 & 62.80 & 99.38 & 2076 & 24.24 & 283.49 & 5978 & 73.33 & 69.03 & 1445 & 59.71 & 118.89 & 2499 \\
No Reasoning & 80.51 & 12.72 & 267 & 58.33 & 15.15 & 315 & 32.83 & 28.16 & 595 & 63.03 & 12.68 & 263 & 58.68 & 17.18 & 360 \\
\cmidrule[0.3pt]{1-16}
Reasoning Ratio($r=0.25$) & 76.63 & 15.52 & 340 & 51.20 & 36.58 & 768 & 19.80 & 97.74 & 2146 & 66.13 & 27.98 & 588 & 53.44 & 44.46 & 963 \\
Reasoning Ratio($r=0.5$) & 77.04 & 18.49 & 403 & 53.90 & 58.99 & 1236 & 20.61 & 159.96 & 3455 & 70.86 & 40.87 & 850 & 55.60 & 69.58 & 1525 \\
Reasoning Ratio($r=0.75$) & 77.45 & 21.16 & 486 & 59.10 & 79.42 & 1731 & 25.05 & 220.97 & 4885 & 73.12 & 54.95 & 1195 & 58.68 & 94.13 & 2049 \\
\cmidrule[0.3pt]{1-16}
SFT Model  & 63.99 & 3.99 & 84 & 46.45 & 10.53 & 220 & 24.75 & 23.51 & 496 & 49.29 & 17.12 & 358 & 46.12 & 13.79 & 289 \\
Answer Convergence & 78.18 & 15.33 & 385 & 59.13 & 54.93 & 538 & 28.98 & 174.58 & 3959 & 65.38 & 43.78 & 840 & 57.92 & 72.16 & 1431 \\
\rowcolor{pale_green} \textbf{SyncThink} & 78.32 & 19.85 & 419 & 59.98 & 41.91 & 872 & \textbf{33.84} & \textbf{76.29} & \textbf{1602} & 67.68 & 38.24 & 796 & 59.96 & 44.07 & 922 \\
\bottomrule
\end{tabular}%
}
\caption{Comparison of different methods across GSM8K, MMLU, GPQA, and BBH datasets on DeepSeek-R1-Qwen2.5 and LLaMA3.1 models. We report Top@1 accuracy, Time, and Tokens usage.}
\label{tab:main_results}
\end{table*}